\documentclass[10pt,twocolumn,letterpaper]{article}

\usepackage{iccv,float}
\usepackage{times}
\usepackage{epsfig}
\usepackage{graphicx}
\usepackage{amsmath}
\usepackage{amssymb}
\usepackage{courier}
\usepackage{color,colortbl}
\usepackage[linesnumbered,lined,boxed,commentsnumbered]{algorithm2e}
\definecolor{Gray}{gray}{0.87}
\usepackage[pagebackref=true,breaklinks=true,letterpaper=true,colorlinks,bookmarks=false]{hyperref}

\iccvfinalcopy 
\def\iccvPaperID{000} 

\ificcvfinal\fi
\begin{document}

\title{Restoration of Non-rigidly Distorted Underwater Images using a Combination of Compressive Sensing and Local Polynomial Image Representations}

\author{Jerin Geo James\\
IIT Bombay\\
{\tt\small jeringeo@cse.iitb.ac.in}
\and
Pranay Agrawal\\
IIT Bombay\\
{\tt\small pranay.agr09@gmail.com}
\and
Ajit Rajwade\\
IIT Bombay\\
{\tt\small ajitvr@cse.iitb.ac.in}
}

\maketitle

\begin{abstract}
Images of static scenes submerged beneath a wavy water surface exhibit severe non-rigid distortions. The physics of water flow suggests that water surfaces possess spatio-temporal smoothness and temporal periodicity. Hence they possess a sparse representation in the 3D discrete Fourier (DFT) basis. Motivated by this, we pose the task of restoration of such video sequences as a compressed sensing (CS) problem. We begin by tracking a few salient feature points across the frames of a video sequence of the submerged scene. Using these point trajectories, we show that the motion fields at all other (non-tracked) points can be effectively estimated using a typical CS solver. This by itself is a novel contribution in the field of non-rigid motion estimation. We show that this method outperforms state of the art algorithms for underwater image restoration. We further consider a simple optical flow algorithm based on local polynomial expansion of the image frames (PEOF). Surprisingly, we demonstrate that PEOF is more efficient and often outperforms all the state of the art methods in terms of numerical measures. Finally, we demonstrate that a two-stage approach consisting of the CS step followed by PEOF much more accurately preserves the image structure and improves the (visual as well as numerical) video quality as compared to just the PEOF stage. 
\end{abstract}

\section{Introduction}
Underwater image analysis is a challenging and relatively less explored area of computer vision. In particular, if a scene submerged in water is imaged by a camera in air, the scene exhibits severe spatial distortions due to dynamic refraction at the wavy/dynamic water surface. Such distortions can interfere with higher level tasks such as object recognition, tracking, motion analysis or segmentation, which are required in applications such as coral reef monitoring, surveillance of shallow riverbeds to observe vegetation \cite{Turlaev2013}, or studying of visual perception in water-birds (see references in \cite{Alterman2013}). Applying the principle of reversibility of light, there also exist applications in submarines where a camera in water is observing scenes in the air \cite{Alterman2013}.

\textbf{Related Work:} There exists a medium-sized body of literature on this topic. The earliest work, to our knowledge, is from \cite{Murase1992} where \emph{frame-to-frame} optical flow is estimated using a correlation-based method, and the underlying image is estimated from the centroid of the flow trajectory at each point. Such a method is expensive and error-prone due to ambiguities in optical flow (especially in case of large motion), and reflective or blur artifacts. The work in \cite{Tian2009} infers a set of `water bases' from synthetic underwater scenes generated from the wave equation, and then expresses deformation fields within small patches as linear combinations of the water bases. The work in \cite{Oreifej2011} performs non-rigid registration of blurred versions of all frames in the sequence with an evolving template (initialized to be the mean image) followed by a robust PCA step \cite{Candes2011}. Both these methods are  expensive and prone to local minima in case of large motion, leaving behind some residual motion or geometric distortion. Along similar lines, \cite{Halder2017} proposes a method to register all frames of a video sequence with a `reference frame', chosen to be a frame with the least blur. The so-called `lucky region approach' has been developed in \cite{Efros2004}, \cite{Donate2007}, \cite{Wen2010} and \cite{Wen2009}. In this, distortion-free patches which correspond to patches of the image formed due to a locally flat portion of the water surface, are identified and then stitched together using graph embedding. In \cite{Seemakurthy2015}, the restoration is performed based on the assumption that the water surface is a single unidirectional cyclic wave (UCW). The restoration process is framed as a blind deconvolution problem, the input to which is the average of the video frames (equivalent to a motion-blurred image due to long camera exposure). The UCW assumption will not hold in some scenarios, for example if the water waves are formed from a superposition of different constituent waves due to multiple independent disturbances. Recently, \cite{Li2018} presented a deep learning framework to restore \emph{single} distorted underwater images (as opposed to using video sequences). A neural network was trained on pairs of distortion-free and distorted images, to infer geometric distortion and also apply photometric correction. This method does not account for the extra information in the form of temporal redundancy, which is readily available in even short video sequences. To be effective, it also requires a large amount of training data. In contrast, our method is based on principles of basic physics/geometry. It also does not require training data as in \cite{Li2018}, representative distortion-free templates (which are generally hard to acquire) to drive the technique as in \cite{Tian2012,Tian2015}, multiple illumination sources as in \cite{Turlaev2013} or multiple viewpoints as in \cite{Qian2018}. 

\textbf{Overview:} In this paper, we present a novel method that exploits the inherent spatio-temporal redundancy of water waves. We note that the motion vector fields (MVFs), defined as the collection of displacement vectors at each point in every frame of the video sequence, have a sparse representation in the discrete Fourier basis. This emerges from the spatial smoothness of the flow, and its temporal periodicity as well as smoothness. We begin by tracking some $N$ salient feature points across all $T$ frames of the video, to yield point-trajectories (PTs) $\{\boldsymbol{p_i} \triangleq \{(x_{it},y_{it})\}_{t=1}^T\}_{i=1}^N$, and then convert these into displacement-trajectories (DTs) $\{\boldsymbol{d_i} \triangleq \{(d_{x,it},d_{y,it})\}_{t=1}^T\}_{i=1}^N$. Given these DTs, we use a compressed sensing (CS) method to infer the DTs at all other points in the image domain. The specific manner in which we have applied CS for motion estimation in this paper, is a novel contribution (see Sec. \ref{subsubsec:comments} for a comparison to other approaches for sparsity-based motion estimation). We also observe that our method largely reduces the non-rigid motion and outperforms the state of the art methods. Our second major contribution is the usage of an existing optical flow method based on local polynomial image representations \cite{Farneback2003}, for this particular task. Despite its simplicity, we show that this method outperforms the state of the art. Lastly, we show that a two stage approach with CS followed by the optical flow method leads to better video stabilization as well as image structure preservation (visual and numerical) than the optical flow stage alone, at only slightly greater computational cost.  

\textbf{Organization:} The main theory behind our method is explained in Sec. \ref{sec:theory}. The datasets and experiments are described in Sec. \ref{sec:exp}, followed by a discussion and conclusion in Sec. \ref{sec:disc}.

\section{Theory}
\label{sec:theory}
We first present a complete description of the various assumptions made in our restoration task, and compare them with those of other aforementioned methods.
\subsection{Image Formation}
We assume a static planar scene submerged at unknown depth $h$ below a clear, shallow water surface, imaged by an orthographic camera in air whose optical axis is pointing downwards perpendicular to the scene. This assumption is valid in practice (as also seen from our results on real sequences in Section \ref{sec:exp}) and has also been made in existing work such as \cite{Tian2009,Seemakurthy2015,Murase1992}. Let $I_0$ be the original image (size $N_x \times N_y$) as if it were formed in the absence of any wavy water surface. Then the distorted image $I_d$ due to the wavy water surface is given as
$I_d(x_0,y_0,t) = I_0(x_0 + d_x(x_0,y_0,t),y_0 + d_y(x_0,y_0,t))$, where $(d_x(x_0,y_0,t),d_y(x_0,y_0,t))$ is the displacement at point $(x_0,y_0)$ (indexing into the \emph{undistorted} image $I_0$) at time $t$. A precise relationship between $(d_x(x_0,y_0,t),d_y(x_0,y_0,t))$ and the derivatives of the dynamic height $z(x,y,t)$ of the water surface at the point of refraction, has been derived in previous work \cite{Murase1992}. 
Here, our aim is to estimate $I_0(x_0,y_0)$ for all $(x_0,y_0)$ given $\{I_d(:,:,t)\}_{t=1}^T$. We are assuming that the video frames are largely blur-free, though moderate deviations from this assumption do not affect our theory or results. We ignore effects such as reflection of light from the water surface (which were found to be absent or rare in real videos we gathered or those from \cite{Tian2009}). 
\\
\subsection{Water Surface and Motion Vector Field Model} \label{subsec:model}
In our work, we essentially require the wavy water surface to be a smooth signal in space and time, and also temporally periodic. The assumption of spatio-temporal smoothness is common in the literature on this topic, for example \cite{Tian2009, Oreifej2011}, and excludes turbulent flows. We do not require the water surface to be explicitly modelled as a linear combination of sinusoidal waves, though our method works very well even for such cases. The motion vector at point $(x_{i0},y_{i0})$ (of the underlying undistorted video) at time instant $t$ is denoted as $\boldsymbol{d^{(t)}_i} \triangleq (d_{xit},d_{yit})$. The complete motion vector field (MVF) can be represented as two 3D signals $\boldsymbol{d_x}, \boldsymbol{d_y}$ in $\mathbb{R}^{N_x \times N_y \times T}$, containing the $X$- and $Y$-components of the displacements at every pixel $(x_{i0},y_{i0})$ and every time instant $t$. Due to the spatio-temporal smoothness (and thereby their band-limited nature), both $\boldsymbol{d_x}, \boldsymbol{d_y}$ will admit a sparse (or compressible) decomposition in the Fourier space. For computational reasons, we use the \emph{Discrete} Fourier Transform (DFT) basis. Given the innate interdependence between $\boldsymbol{d_x}, \boldsymbol{d_y}$ (since they emerge from the same wavy water surface), we instead work with a complex-valued vector field $\boldsymbol{d} \triangleq \boldsymbol{d_x} + \iota \boldsymbol{d_y}$ where $\boldsymbol{d} \in \mathbb{C}^{N_x \times N_y \times T}$ and $\iota \triangleq \sqrt{-1}$. This is a natural way to exploit the interdependence. Moreover, if the video sequence is long enough so that the actual MVF is \emph{temporally periodic}, then that further contributes to the sparsity of the Fourier domain representation. This is because by definition, periodic signals are sparse in the Fourier domain, even more so if they are band-limited - which is a consequence of smoothness. 

The assumption of sparsity of the MVF in the Fourier basis finds corroboration in the fluid mechanics literature. For example in  \cite{Rosenfeld1995,Fenton1982,Willcox2007}, the  rapid convergence of the Fourier series of different types of time-periodic velocity vector fields arising in fluid flow, has been demonstrated. Due to this, the water surface height $z(x,y,t)$ and hence the MVFs (which are related to derivatives of $z(x,y,t)$) will also admit sparse Fourier decomposition. In addition, in Sec. \ref{subsec:discussion}, we present an empirical verification of the Fourier sparsity of the MVF $\boldsymbol{d}$ from real underwater video sequences. 
\subsection{Method Overview}
An overview of our method is summarized in a pseudo-code in Alg. \ref{alg:Alg1}. The detailed steps are described further.
\begin{algorithm}
\SetAlgoLined
\SetKwInOut{Input}{input}\SetKwInOut{Output}{output}
\Input{Distorted video $I_d$}
\Output{Restored image $\bar{I_r}$}
Track $N$ feature points to obtain point-trajectories $\{\boldsymbol{p_i}\}_{i=1}^N$ as per Sec. \ref{subsubsec:tracking}.\\
Compute displacement trajectories $\{\boldsymbol{d_i}\}_{i=1}^N$ as per Sec. \ref{subsubsec:displ}.\\
Compute the motion vector field (MVF) $\boldsymbol{d}$ as defined in Sec. \ref{subsec:model} from its measurements $\{\boldsymbol{d_i}\}_{i=1}^N$ using the CS-based method from Sec. \ref{subsubsec:mvf_cs}.\\
Perform motion correction from the computed MVF to obtain a restored video $I_{r}$. \\
Optionally, perform further motion correction on $I_{r}$ using the PEOF technique from Sec. \ref{subsubsec:PEOF}. \\
Compute mean or median frame of $I_r$ to yield $\bar{I_r}$.
\caption{Algorithm to Restore Video}
\label{alg:Alg1}
\end{algorithm}
\subsubsection{Feature point detection and tracking}
\label{subsubsec:tracking}
We begin with a salient feature point detection and tracking algorithm yielding point-trajectories $\{\boldsymbol{p_i} \triangleq \{(x_{it},y_{it})\}_{t=1}^T\}_{i=1}^N$ for $N$ salient feature points detected in the first frame. The coordinates $(x_{it},y_{it})$ represent the position in frame $t$ of the $i^{\textrm{th}}$ point whose coordinates in a distortion-free frame are denoted as $(x_{i0},y_{i0})$, where the subscript `$0$' refers to an index in the undistorted image. Of course, $(x_{i0},y_{i0})$ are unknown at the outset. Our salient feature point detection combines four algorithms: (1) difference of Gaussians (DoG) used by SURF\cite{Bay2008}, (2) the FAST method \cite{Rosten2010}, (3) the popular Harris corner method, and (4) the BRISK technique \cite{BRISK}. Consider a union-set of salient points in the first frame, as detected by all these methods. All points in this set are tracked in subsequent frames using the well-known Kanade-Lucas-Tomasi (KLT) tracker \cite{MATLAB_KLT}. We obtain excellent results with the standard KLT tracker because it inherently takes care of locally affine motion (a first approximation to non-rigid motion). In some cases, however, we encounter tracking errors. Such trajectories are weeded out and not used in later steps, if (1) they are considered invalid by the KLT tracker itself (which happens when the region around a salient feature point in a frame cannot be accurately expressed as an affine transformation of the corresponding region in a previous frame), or if (2) the center of trajectory (COT), as defined in Sec. \ref{subsubsec:displ}, computed over the first $T/2$ and last $T/2$ frames differ by a threshold of more than 3 pixels. 

We also trained a Siamese network following \cite{Simo-Serra2015} to learn good feature descriptors. See \texttt{supplemental material} for further details about this. The Siamese network produced slightly better results than the KLT tracker on unseen synthetic and real data. However it did not perform as well as the KLT tracker if there was blur in the video frames. Hence we used the KLT tracker in all experiments. Examples of point tracking on real sequences are shown in the supplemental material folder `CS\_MotionReduction'. 
\subsubsection{Displacement computation}
\label{subsubsec:displ}
Following previous definitions of $(x_{i0},y_{i0})$ and the point-trajectory $\boldsymbol{p_i}$, we approximate $\hat{x}_{i0} \triangleq \sum_{t=1}^T x_{it}/T \approx x_{i0}; \hat{y}_{i0} \triangleq \sum_{t=1}^T y_{it}/T \approx y_{i0}$ (termed `center of trajectory' or COT), although more robust `means' such as the median can also be considered. This approximation is well justified by the assumption of the local symmetry of water motion, due to which the average surface normal (across time) at any point on the water surface is close to the vertical line \cite{Murase1992}. Our experiments for synthetic and real sequences confirm that this is valid even for moderate $T \sim 50$ frames. The supplemental material includes an illustrative example. With this, our set of displacements for the $i^{\textrm{th}}$ salient feature point are given as $\boldsymbol{d_i} \triangleq (\boldsymbol{d_{ix}},\boldsymbol{d_{iy}}) \triangleq \{(x_{it}-\hat{x}_{i0},y_{it}-\hat{y}_{i0})\}_{t=1}^T$. We term these as `displacement-trajectories' (DTs).

\subsubsection{MVF Estimation using CS}
\label{subsubsec:mvf_cs}
The DTs $\{\boldsymbol{d_i}\}_{i=1}^N$ can be regarded as sparse samples (in the space-time domain) of the 3D MVF signal $\boldsymbol{d}$. The signal $\boldsymbol{d}$ is sparse in the Fourier domain (see Sec. \ref{subsec:model}) and hence can be expressed as $\textrm{vec}(\boldsymbol{d}) = \boldsymbol{F \theta}$ where $\boldsymbol{F}$ is the 3D-DFT basis matrix and $\boldsymbol{\theta}$ is a sparse vector of Fourier coefficients. If the DTs are concatenated to form a complex-valued `measurement vector' $\boldsymbol{e}$ of $NT$ elements, then we have the following model:
\begin{equation}
\boldsymbol{e} = \boldsymbol{\Phi F \theta} + \boldsymbol{\eta},
\label{eq:CS_MVF}
\end{equation}
where $\boldsymbol{\Phi}$ is a sampling matrix of size $NT \times N_xN_yT$ and $\boldsymbol{\eta}$ is a noise vector of $NT$ elements indicating errors in the DTs obtained from the tracking algorithm. Note that $\boldsymbol{\Phi}$ is the row-subsampled version of the identity matrix of size $N_xN_yT \times N_xN_yT$, and each row of $\boldsymbol{\Phi}$ is a one-hot vector which indicates whether or not the displacement at some pixel $(x_{i0},y_{i0})$ (in the undistorted image) at some time frame $t$ was included in the set $\{\boldsymbol{d_i}\}_{i=1}^N$ (and hence the measurement vector $\boldsymbol{e}$). 
The sensing matrix $\boldsymbol{\Phi}$ and representation matrix $\boldsymbol{F}$ are an ideal combination, because they are highly incoherent with each. This augurs well for the application of a CS algorithm for estimation of $\boldsymbol{\theta}$ (and thereby $\boldsymbol{d}$) from $\boldsymbol{e}, \boldsymbol{\Phi}$. This is because CS theory states that $\mathcal{O}(s \log (N_x N_y T) \mu(\boldsymbol{\Phi F}))$ measurements are sufficient for accurate reconstruction of the $s$-sparse vector $\boldsymbol{\theta}$ with very high probability \cite{CRT}. Here $\mu(\boldsymbol{\Phi F})$ is the coherence between $\boldsymbol{\Phi}$ and $\boldsymbol{F}$ and is defined as $\mu(\boldsymbol{\Phi F}) \triangleq \textrm{max}_{i,j,i \neq j}\frac{|\boldsymbol{\Phi^i F_j}|}{\|\boldsymbol{\Phi^i}\|_2 \|\boldsymbol{F_j}\|_2}$ where $\boldsymbol{\Phi^i}$ and $\boldsymbol{F_j}$ are the $i^{\textrm{th}}$ row of $\boldsymbol{\Phi}$ and $j^{\textrm{th}}$ column of $\boldsymbol{F}$ respectively. Given the choice of $\boldsymbol{\Phi},\boldsymbol{F}$ for our task, $\mu$ reaches its lower bound of 1, thereby reducing the number of samples required for reconstruction guarantees. 
To account for the noise in $\boldsymbol{e}$, we determine $\boldsymbol{d}$ using an estimator (popularly called the LASSO) which minimizes the following objective function:
\begin{equation}
J(\boldsymbol{\theta}) = \lambda \|\boldsymbol{\theta}\|_1 + \|\boldsymbol{e}-\boldsymbol{\Phi F \theta}\|^2.
\label{eq:lasso}
\end{equation}
The regularization parameter $\lambda$ can be chosen by cross-validation \cite{Ward2009} from a set $\mathcal{S}$ of candidate values. That is, for every $\hat{\lambda} \in \mathcal{S}$, a candidate signal $\boldsymbol{d}_{\hat{\lambda}}$ is computed by the LASSO method using a set $\mathcal{T}_1$ of only (say) 90\% of the measurements from $\boldsymbol{e}$. Following this, the value of $E(\boldsymbol{d}_{\hat{\lambda}}) \triangleq \sum_{i \in \mathcal{T}_2}|e_i-\boldsymbol{\Phi^i}\boldsymbol{d}_{\hat{\lambda}}|^2$ is computed, where $\mathcal{T}_2$ is the set of the remaining measurements in $\boldsymbol{e}$. The value of $\hat{\lambda}$ that minimizes $E(\boldsymbol{d}_{\hat{\lambda}})$ can be selected. Following this, $\boldsymbol{d}$ is re-estimated by the LASSO method from \emph{all} measurements in $\boldsymbol{e}$ 
and with the selected $\hat{\lambda}$ value. 

\subsubsection{Comments Regarding MVF Estimation using CS}
\label{subsubsec:comments}
Note that our method is very different from the bispectral approach in \cite{Wen2010} which chooses `lucky' (i.e. least distorted) patches, by comparing to a mean template. In that method, the Fourier transform is computed locally on small patches in the spatial domain for finding similarity with corresponding patches from a mean image. On the other hand, our Fourier decomposition is spatio-temporal and global. The idea of dense optical flow interpolation (not specific to underwater scenes) from a sparse set of feature point correspondences has been proposed in the so-called EpicFlow technique \cite{Revaud2015}. The interpolation uses non-parametric kernel regression or a locally affine method. However our method uses key properties (spatio-temporal smoothness and temporal periodicity) of water waves, and thus considers temporal aspects of the MVFs. This aspect is missing in EpicFlow. Nevertheless, we present comparisons to EpicFlow in Sec. \ref{sec:exp}. 

The use of sparsity based techniques for dense flow field estimation is not entirely new and has been used earlier in \cite{Dong2014,Shen2010,Jia2011}. However besides the usage of sparsity for underwater image restoration, there are key differences between our approach and the existing ones. (a) First, these papers use a sparse representation (\textit{eg.}, wavelets \cite{Shen2010}, learned dictionaries \cite{Jia2011} or low-rank and sparse models \cite{Dong2014}) for optical flow in small patches unlike our method which is more global. (b) Second, they compute the optical flow only between two frames with a data fidelity term based on the brightness constancy equation (unlike our approach which uses displacement trajectories), they do not consider spatio-temporal patches, and do not account for \emph{temporal redundancy}, which is a readily available and useful prior that our approach exploits. 

\subsubsection{Polynomial Image Expansions for Optical Flow}
\label{subsubsec:PEOF}
The classical optical flow method in \cite{Farneback2003} expresses small patches from the two images $f_1$ and $f_2$, between which the MVF has to be computed, as second-degree polynomials. This method can unduly smooth motion discontinuities as mentioned in \cite{Farneback2003}, but it is well suited to our problem, due to the spatial smoothness of water waves. Consider the following:
\begin{align}
f_1(\boldsymbol{x}) &= \boldsymbol{x}^t \boldsymbol{A_1} \boldsymbol{x} + \boldsymbol{b_1}^t \boldsymbol{x} + \boldsymbol{c_1} \\ \nonumber
f_2(\boldsymbol{x}) &\approx f_1(\boldsymbol{x}-\boldsymbol{d})  = (\boldsymbol{x}-\boldsymbol{d})^t \boldsymbol{A_1} (\boldsymbol{x}-\boldsymbol{d}) + \boldsymbol{b_1}^t (\boldsymbol{x}-\boldsymbol{d}) + \boldsymbol{c_1} \\ \nonumber
&\approx \boldsymbol{x}^t \boldsymbol{A_2} \boldsymbol{x} + \boldsymbol{b_2}^t \boldsymbol{x} + \boldsymbol{c_2},
\end{align}
where $\boldsymbol{d}$ is the 2D displacement vector at the point $\boldsymbol{x} \triangleq (x,y)^t$. Consider small patches in the two images respectively, centered around point $\boldsymbol{x}$. The polynomial coefficients $\boldsymbol{A_1},\boldsymbol{A_2} \in \mathbb{R}^{2 \times 2}; \boldsymbol{b_1},\boldsymbol{b_2},\boldsymbol{c_1},\boldsymbol{c_2} \in \mathbb{R}^{2 \times 1}$ can be determined by local regression. This process is repeated in sliding window fashion all through the image, and so these coefficients become functions of $\boldsymbol{x}$. Assuming a slowly changing MVF, the displacement $\boldsymbol{d}(\boldsymbol{x})$ can be computed in the following manner:
\begin{equation}
\boldsymbol{d}(\boldsymbol{x}) = \Big(\sum_{\boldsymbol{\tilde{x}} \in \mathcal{N}(\boldsymbol{x})} \boldsymbol{A}(\boldsymbol{\tilde{x}})^T \boldsymbol{A}(\boldsymbol{\tilde{x}})\Big)^{-1} \sum_{\boldsymbol{\tilde{x}} \in \mathcal{N}(\boldsymbol{x})}  \boldsymbol{A}(\boldsymbol{\tilde{x}})^T \boldsymbol{\Delta b(\boldsymbol{\tilde{x}})},
\end{equation}
where $\boldsymbol{\tilde{x}}$ is a point in a neighborhood $\mathcal{N}(\boldsymbol{x})$ around $\boldsymbol{x}$, $\boldsymbol{A}(\boldsymbol{\tilde{x}}) \triangleq (\boldsymbol{A_1}(\boldsymbol{\tilde{x}})+\boldsymbol{A_2}(\boldsymbol{\tilde{x}}))/2$ and $\boldsymbol{\Delta b(\boldsymbol{\tilde{x}})} \triangleq (\boldsymbol{b_1}(\boldsymbol{\tilde{x}})-\boldsymbol{b_2}(\boldsymbol{\tilde{x}}))/2$. Further details about this method can be found in \cite{Farneback2003}. We term this method as polynomial expansion based optical flow (PEOF). Given $\boldsymbol{d}(\boldsymbol{x})$, the image $f_1$ is warped, followed by polynomial fitting in the warped version of $f_1$, and re-estimation of $\boldsymbol{d}(\boldsymbol{x})$.  The procedure is repeated iteratively. In the present work, the PEOF method is used to find the MVF between each video frame and the mean image (the average of all video frames). The computed MVFs are applied to each frame to obtain the restored video. These restored images are then averaged to yield a final restored image. As shown in Sec. \ref{sec:exp}, this method outperforms all state of the art methods in terms of image restoration quality as well as computation speed. 
\section{Experimental Results}
\label{sec:exp}
In this section, we present an extensive suite of results on both synthetic and real video sequences. All image and video results are available in the \texttt{supplemental material}. 

\subsection{Description of Datasets}
We created several synthetic 50 fps videos of size $\sim 512 \times 512 \times 101$ by simulating the refraction model from \cite{Murase1992} on different images containing objects/text, for a scene depth of 25 cm below the water surface. The water surface was generated using superposition of $2 \leq K \leq 6$  sinusoidal waves with randomly chosen parameters. We henceforth refer to this dataset as \textbf{Synthetic}. We also gathered real video sequences (of size $\sim700 \times 512 \times 101$ with a 50 fps camera) of laminated posters kept at the bottom of a water-tank, with waves generated by mechanical paddles. See \texttt{supplemental material} for details of the acquisition. Visual inspection revealed that blur was occasionally present in some frames. We henceforth refer to this dataset as \textbf{Real1}. For ground truth ($I_0$), we also acquired a single image of the same posters under still water with the same camera settings. We also demonstrate results on three text sequences (size $\sim 300 \times 250 \times 101$ at 125 fps) obtained from \cite{Tian2009}, for which ground truth was available. We henceforth refer to this dataset as \textbf{Real2}. Note that Real1 is a more challenging dataset than Real2 due to greater frame-to-frame motion - see Table \ref{tab:motion_reduction} for the standard deviation values of the motion $\sigma_{motion} \triangleq \sqrt{\sum_{i=1}^N \sum_{t=1}^T((x_{it}-\hat{x_{i0}})^2 + (y_{it}-\hat{y_{i0}})^2)/(NT-1)}$, computed over salient point trajectories.

\subsection{Key Parameters and Comparisons}
We compared restoration results for several algorithms: (1) our CS-based method (CS) from Sec. \ref{subsubsec:mvf_cs}; (2) our PEOF method from Sec. \ref{subsubsec:PEOF}; (3) the CS-based method followed by the PEOF method (CS+PEOF); (4) the two-stage method in \cite{Oreifej2011} consisting of spline-based registration followed by RPCA (SBR-RPCA) which is considered state of the art for underwater image restoration; (5) the method from \cite{Tian2009} using learned water bases (LWB); and (6) the deep learning (DL) approach from \cite{Li2018}. For SBR-RPCA and LWB, we used code provided by the authors with default parameters. For DL, we used the pre-trained network and code provided by the authors, on each video frame separately and then computed the mean image. We performed all computation and quality assessment with each video frame resized to $256 \times 256$ (after suitable cropping to maintain aspect-ratio), as required by their specific implementation. For CS, we used the well-known YALL1 (Basic) solver \cite{YALL1}, which allows for $\ell_1$-norm optimization of complex-valued signals. We observed better and faster results in practice by downsampling the DTs comprising $\boldsymbol{e}$ by a factor of 8 in $X,Y$ directions (which is in tune with the bandlimited nature of water waves), followed by CS reconstruction and subsequent upsampling to obtain the final reconstructed MVF. For PEOF, we used the OpenCV implementation with a multi-scale pyramidal approach with 3 levels, a pyramid scale of 0.5 and 10 iterations (i.e. the default parameters). 
For quality assessment referring to ground truth, we used the following measures: (i) visual inspection of the restored video $I_r$ as well as the mean-frame $\bar{I_r}$ of the restored video, (ii) RMSE computed as $\|\bar{I_r}-I_0\|_2/\|I_0\|_2$ where $I_0$ is the image representing the undistorted static scene, (iii) normalized mutual information (NMI) between $\bar{I_r}$ and $I_0$, and (iv) SSIM \cite{Wang2009} between $\bar{I_r}$ and $I_0$. 

We also considered comparing PEOF to a very competitive optical flow algorithm: EpicFlow \cite{Revaud2015} (EF). For EF \cite{Revaud2015}, we used the authors' code to estimate the deformation of all video frames w.r.t. the mean frame of the video. We then applied the  deformations to each frame to yield the final image. Results comparing PEOF and EF are included in the \texttt{supplemental material}, and show that PEOF outperforms EF for underwater image restoration. Note that the EF method has so far \emph{not} been applied for this task in the literature. We do not present results with the state of the art deep learning approaches for optical flow such as \cite{Dosovitskiy2015} or \cite{Sun2018} here, for two reasons: (i) EF yielded superior results on our data compared to \cite{Dosovitskiy2015}, and (ii) the results of PWC-net from \cite{Sun2018} show only a small improvement over EpicFlow on some datasets such as Sintel. 
 
We did not compare with the work in \cite{Seemakurthy2015} because it relies on a unidirectional wave motion assumption (whereas we assume more general wave models), and due to unavailability of publicly released code. 
Also, we did not explicitly compare our results with the method in \cite{Halder2017} for which publicly released code is unavailable. However, we observed that CS-PEOF outperformed the method of \cite{Halder2017} on Real2 (compare `Middle', `Small' and `Tiny' in Table \ref{tab:results} to `Large', `Medium' and `Small' respectively in Table 1 of \cite{Halder2017}).
 
 \begin{figure}
    \centering
    \includegraphics[width=0.4\textwidth]{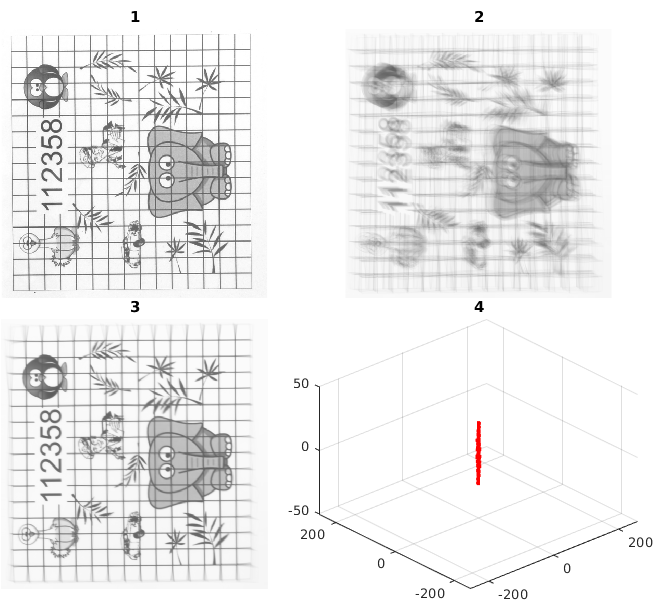}
    \caption{Verification of Fourier domain sparsity of MVF $\boldsymbol{d}$ estimated from a real sequence. Top row: original undistorted image acquired in still water (left), mean of distorted video sequence (right); Bottom row: mean of restored video sequence using CS (left), scatter plot of frequencies which account for 99\% of the squared magnitude of the estimated MVF using CS (right).}
    \label{fig:DT_fourier_sparsity}
\end{figure}
\begin{figure}
\begin{tabular}{l}
\hspace{-0.4in}\includegraphics[width=0.5\textwidth]{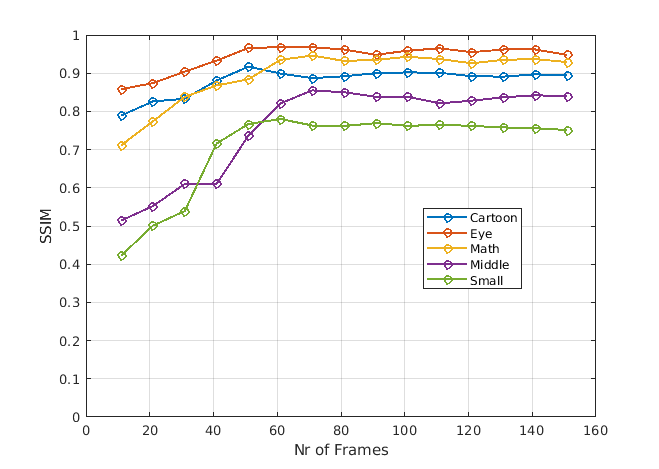}\\
\hspace{-0.4in}\includegraphics[width=0.5\textwidth]{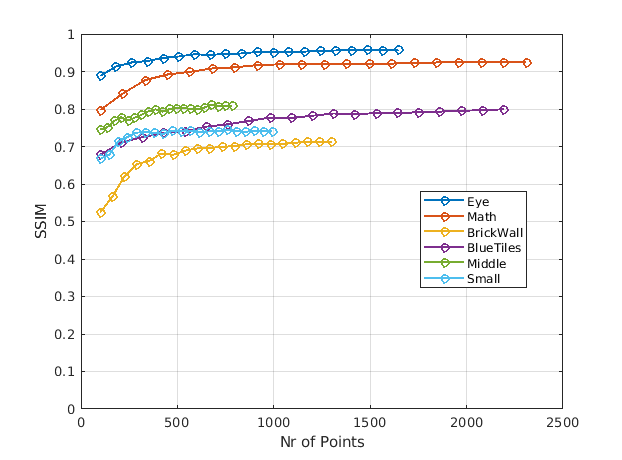}
\end{tabular}
    \caption{Effect of increase in number of frames $T$ (top) and number of salient points $N$ (bottom) on restoration performance for CS method. Results shown on `Middle' and `Small' from Real2, and a few sequences from Real1/Synthetic.}
    \label{fig:NT}
\end{figure}
 
\subsection{Discussion of Results}
\label{subsec:discussion}
\begin{figure}[ht]
\begin{tabular}{lll}
\includegraphics[width=0.3\linewidth]{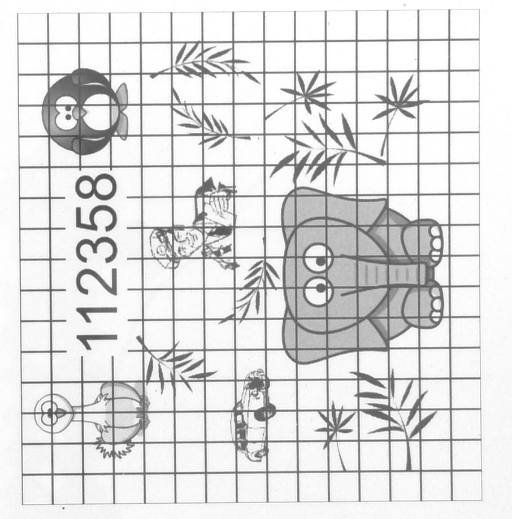}
&\includegraphics[width=0.3\linewidth]{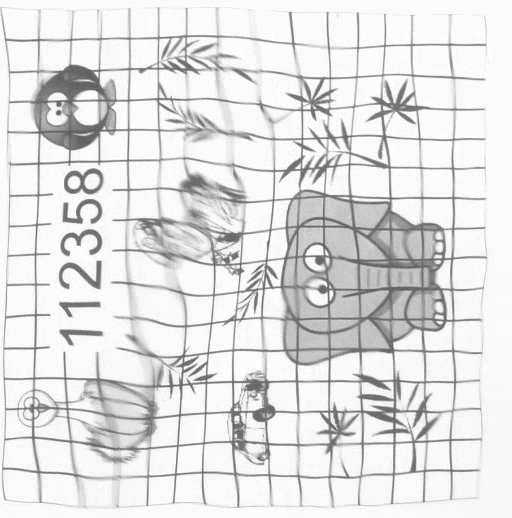} \\
\includegraphics[width=0.3\linewidth]{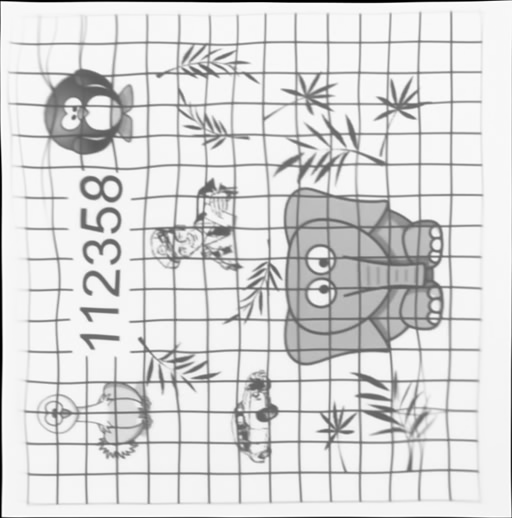}
& \includegraphics[width=0.3\linewidth]{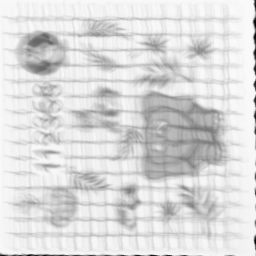}
&\includegraphics[width=0.3\linewidth]{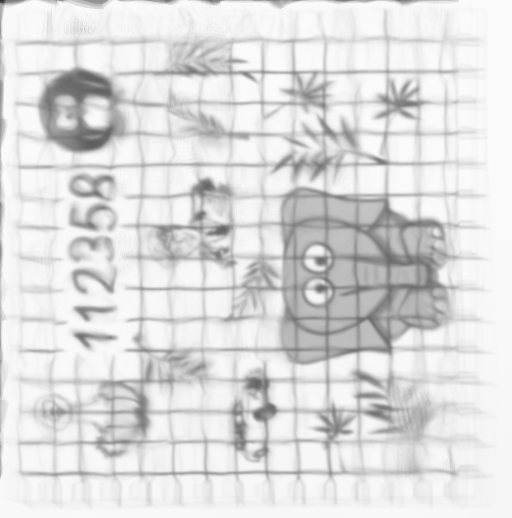}\\
\includegraphics[width=0.3\linewidth]{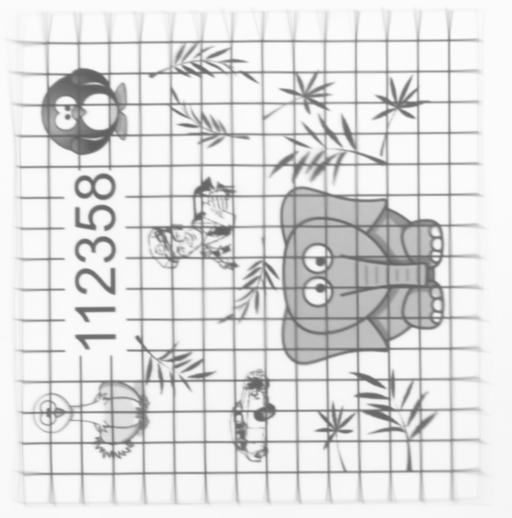}
& \includegraphics[width=0.3\linewidth]{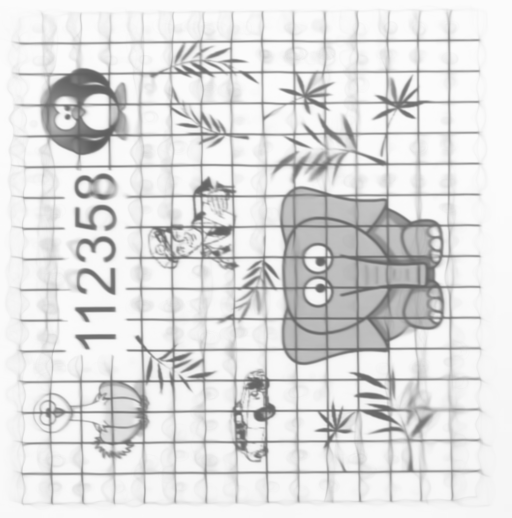}
&\includegraphics[width=0.3\linewidth]{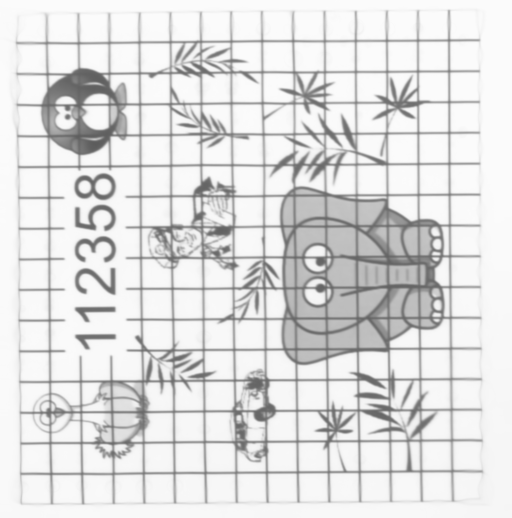} \\
\includegraphics[width=0.3\linewidth]{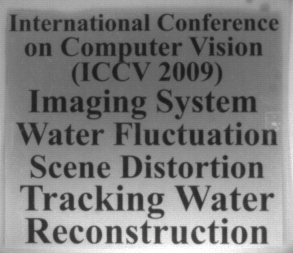} 
&\includegraphics[width=0.3\linewidth]{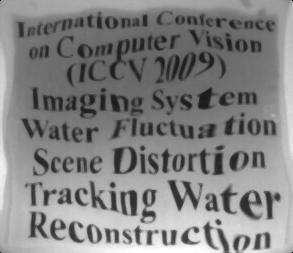} \\
\includegraphics[width=0.3\linewidth]{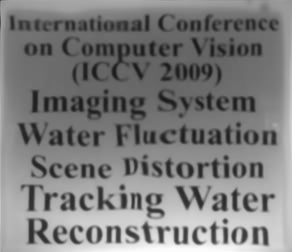}
& \includegraphics[width=0.3\linewidth]{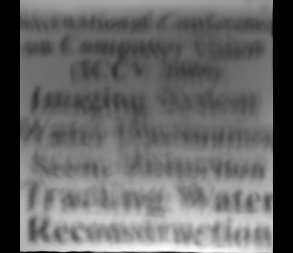} 
&\includegraphics[width=0.3\linewidth]{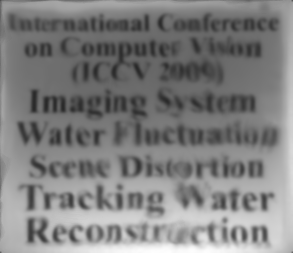} \\
\includegraphics[width=0.3\linewidth]{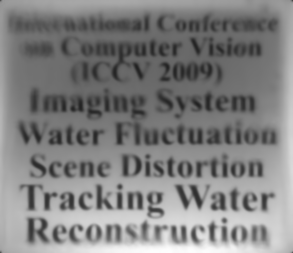}
& \includegraphics[width=0.3\linewidth]{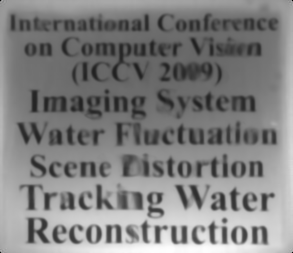}
&\includegraphics[width=0.3\linewidth]{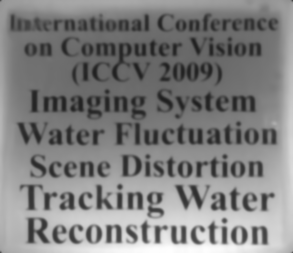}\\
\includegraphics[width=0.3\linewidth]{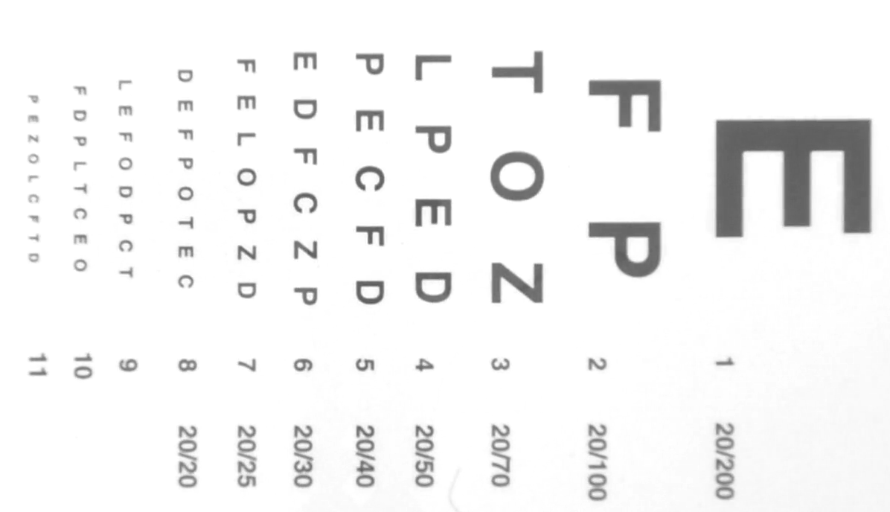} &\includegraphics[width=0.3\linewidth]{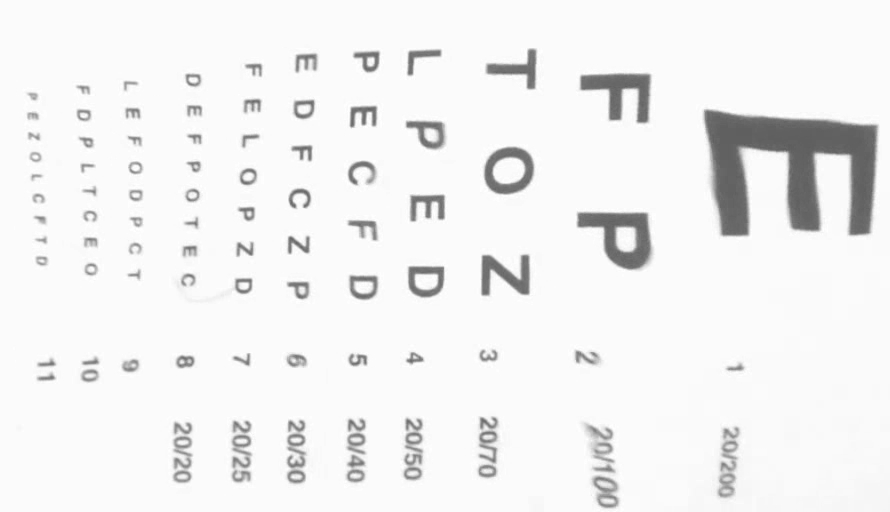} \\
\includegraphics[width=0.3\linewidth]{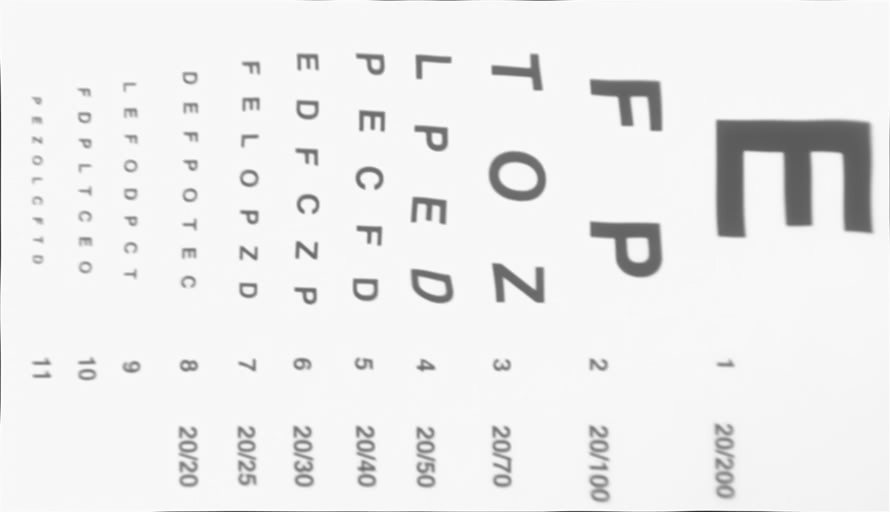}
&\includegraphics[width=0.3\linewidth]{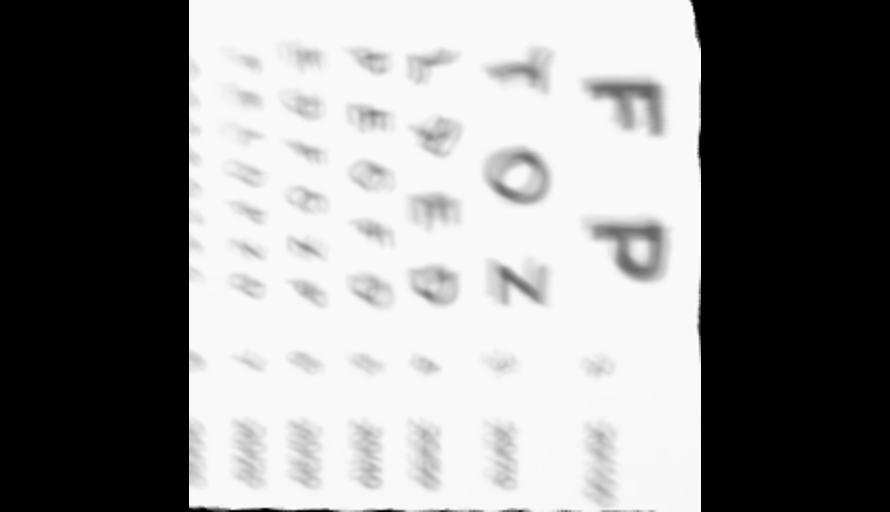} 
&\includegraphics[width=0.3\linewidth]{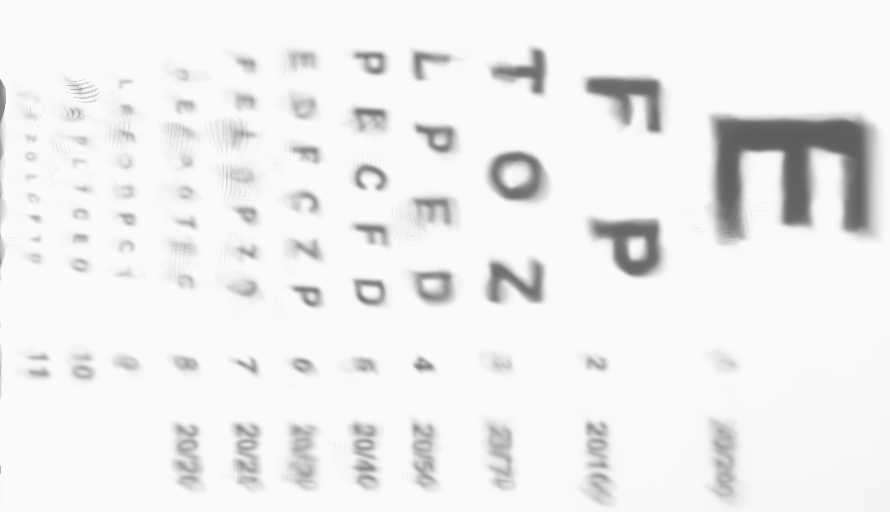} \\
\includegraphics[width=0.3\linewidth]{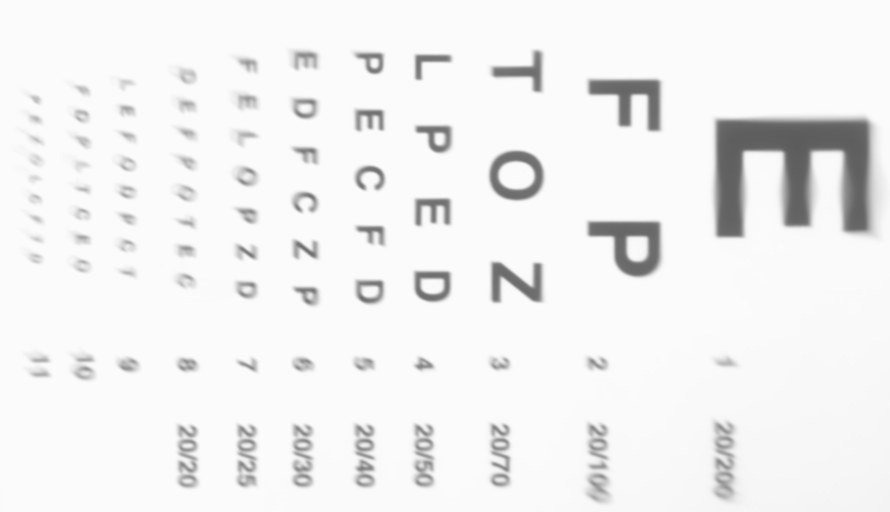}
& \includegraphics[width=0.3\linewidth]{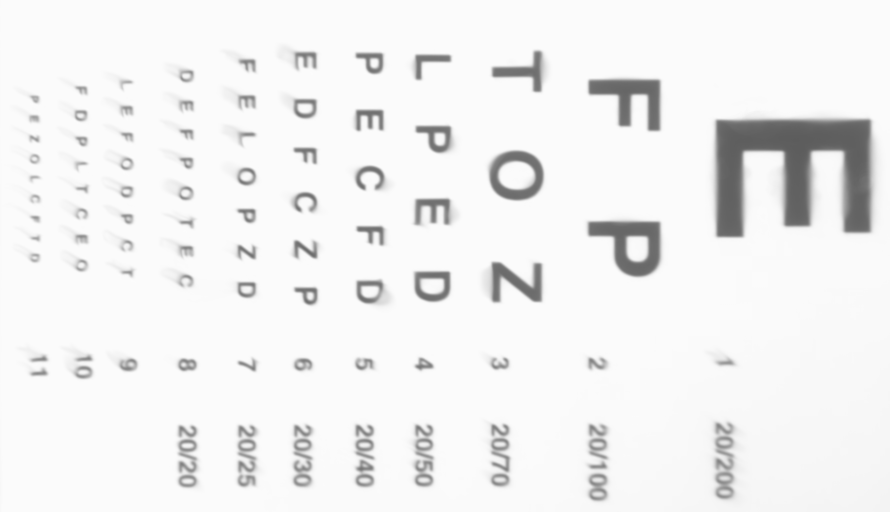}
&\includegraphics[width=0.3\linewidth]{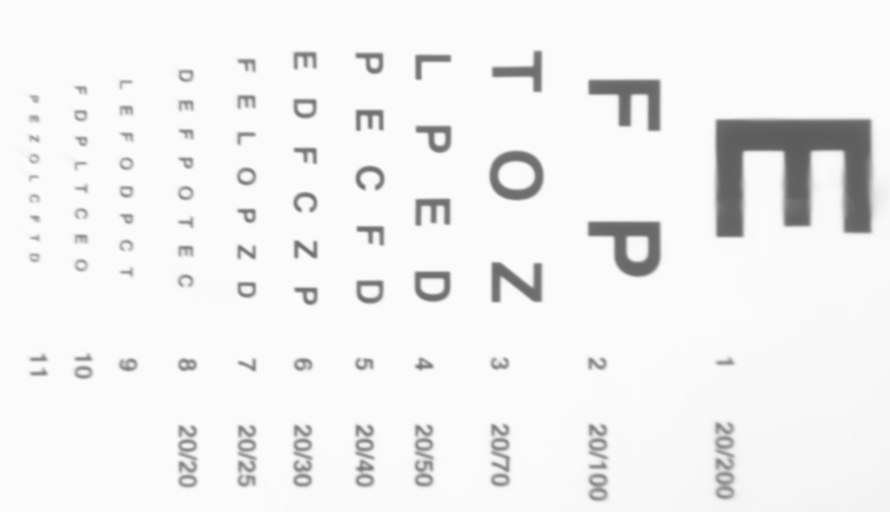}

\end{tabular}
\caption{Left to right, top to bottom order in each of the 3 groups of images: ground truth, distorted sample frame; mean frame restored by SBR-RPCA \cite{Oreifej2011}, DL \cite{Li2018}, LWB \cite{Tian2009}; and by CS, PEOF, CS-PEOF. Zoom into pdf for better view. See \texttt{supplemental material} for more results. Notice geometric distortions in other methods unlike with our methods. The three groups are for `Elephant' (Real1), `Middle' (Real2) and `Eye' (Real1)}
\label{fig:comparison_results}
\end{figure}

The numerical results are presented in Table \ref{tab:results}. The mean images of three real videos restored by various methods are presented in Figs. \ref{fig:comparison_results}. The \texttt{supplemental material} contains results on 14 videos (mean images and restored videos) for all methods. From these results, it is clear that our methods (CS, PEOF and CS-PEOF) yield results surpassing SBR-RPCA, LWB and DL on synthetic as well as real datasets, numerically and also in terms of visual quality. We also supplemented our method with a step involving RPCA \cite{Candes2011} to remove sparse artifacts, which improved video stability but had very little impact over the quality of the mean image. Although PEOF produces superior numerical results to CS, we have observed that CS produces restored videos and mean images with superior  visual quality as compared to PEOF - see Fig. \ref{fig:comparison_results} (grid lines on `Elephant', words `Imaging', `Distortion' in `Middle') as well as \texttt{supplemental material}. 
\begin{table*}[]
\caption{Comparison of various methods on synthetic and real video sequences w.r.t. compute time (h=hours,m=mins.,s=secs.), NMI, SSIM, RMSE. Lower RMSE, higher SSIM and NMI are better.}
\begin{tabular}{lcccccccccccc}
\hline
\multicolumn{1}{|l|}{\textbf{}} & \multicolumn{4}{c|}{\textbf{CS}} & \multicolumn{4}{c|}{\textbf{PEOF}} & \multicolumn{4}{c|}{\textbf{CS+PEOF}} \\ \hline
\multicolumn{1}{|l|}{\textbf{Dataset}} & \multicolumn{1}{c|}{\textbf{Time}} & \multicolumn{1}{c|}{\textbf{NMI}} & \multicolumn{1}{c|}{\textbf{SSIM}} & \multicolumn{1}{c|}{\textbf{RMSE}} & \multicolumn{1}{c|}{\textbf{Time}} & \multicolumn{1}{c|}{\textbf{NMI}} & \multicolumn{1}{c|}{\textbf{SSIM}} & \multicolumn{1}{c|}{\textbf{RMSE}} & \multicolumn{1}{c|}{\textbf{Time}} & \multicolumn{1}{c|}{\textbf{NMI}} & \multicolumn{1}{c|}{\textbf{SSIM}} & \multicolumn{1}{c|}{\textbf{RMSE}} \\ \hline
\rowcolor{Gray}
\multicolumn{1}{|l}{\textit{\textbf{Real1}}} & \multicolumn{1}{c}{} & \multicolumn{1}{c}{} & \multicolumn{1}{c}{} & \multicolumn{1}{c}{} & \multicolumn{1}{c}{} & \multicolumn{1}{c}{} & \multicolumn{1}{c}{} & \multicolumn{1}{c}{} & \multicolumn{1}{c}{} & \multicolumn{1}{c}{} & \multicolumn{1}{c}{} & \multicolumn{1}{c|}{} \\ \hline
\multicolumn{1}{|l|}{Cartoon} & \multicolumn{1}{c|}{0m 42s} & \multicolumn{1}{c|}{1.227} & \multicolumn{1}{c|}{0.902} & \multicolumn{1}{c|}{0.065} & \multicolumn{1}{c|}{0m 41s} & \multicolumn{1}{c|}{1.216} & \multicolumn{1}{c|}{0.913} & \multicolumn{1}{c|}{0.062} & \multicolumn{1}{c|}{1m 23s} & \multicolumn{1}{c|}{1.255} & \multicolumn{1}{c|}{\textbf{0.928}} & \multicolumn{1}{c|}{0.057} \\ \hline
\multicolumn{1}{|l|}{Checker} & \multicolumn{1}{c|}{1m 9s} & \multicolumn{1}{c|}{1.206} & \multicolumn{1}{c|}{0.884} & \multicolumn{1}{c|}{0.104} & \multicolumn{1}{c|}{0m 40s} & \multicolumn{1}{c|}{1.196} & \multicolumn{1}{c|}{0.89} & \multicolumn{1}{c|}{0.105} & \multicolumn{1}{c|}{1m 49s} & \multicolumn{1}{c|}{1.22} & \multicolumn{1}{c|}{\textbf{0.892}} & \multicolumn{1}{c|}{0.104} \\ \hline
\multicolumn{1}{|l|}{Dices} & \multicolumn{1}{c|}{1m 20s} & \multicolumn{1}{c|}{1.172} & \multicolumn{1}{c|}{0.937} & \multicolumn{1}{c|}{0.067} & \multicolumn{1}{c|}{0m 40s} & \multicolumn{1}{c|}{1.139} & \multicolumn{1}{c|}{0.905} & \multicolumn{1}{c|}{0.075} & \multicolumn{1}{c|}{2m 1s} & \multicolumn{1}{c|}{1.188} & \multicolumn{1}{c|}{\textbf{0.956}} & \multicolumn{1}{c|}{0.059} \\ \hline
\multicolumn{1}{|l|}{Bricks} & \multicolumn{1}{c|}{1m 1s} & \multicolumn{1}{c|}{1.148} & \multicolumn{1}{c|}{0.785} & \multicolumn{1}{c|}{0.142} & \multicolumn{1}{c|}{0m 34s} & \multicolumn{1}{c|}{1.151} & \multicolumn{1}{c|}{0.803} & \multicolumn{1}{c|}{0.121} & \multicolumn{1}{c|}{1m 36s} & \multicolumn{1}{c|}{1.167} & \multicolumn{1}{c|}{\textbf{0.843}} & \multicolumn{1}{c|}{0.118} \\ \hline
\multicolumn{1}{|l|}{Elephant} & \multicolumn{1}{c|}{0m 28s} & \multicolumn{1}{c|}{1.128} & \multicolumn{1}{c|}{0.801} & \multicolumn{1}{c|}{0.141} & \multicolumn{1}{c|}{0m 26s} & \multicolumn{1}{c|}{1.102} & \multicolumn{1}{c|}{0.763} & \multicolumn{1}{c|}{0.152} & \multicolumn{1}{c|}{0m 55s} & \multicolumn{1}{c|}{1.132} & \multicolumn{1}{c|}{\textbf{0.808}} & \multicolumn{1}{c|}{0.143} \\ \hline
\multicolumn{1}{|l|}{Eye} & \multicolumn{1}{c|}{1m 22s} & \multicolumn{1}{c|}{1.266} & \multicolumn{1}{c|}{0.961} & \multicolumn{1}{c|}{0.052} & \multicolumn{1}{c|}{0m 57s} & \multicolumn{1}{c|}{1.26} & \multicolumn{1}{c|}{0.975} & \multicolumn{1}{c|}{0.042} & \multicolumn{1}{c|}{2m 19s} & \multicolumn{1}{c|}{1.303} & \multicolumn{1}{c|}{\textbf{0.982}} & \multicolumn{1}{c|}{0.037} \\ \hline
\multicolumn{1}{|l|}{Math} & \multicolumn{1}{c|}{1m 19s} & \multicolumn{1}{c|}{1.193} & \multicolumn{1}{c|}{0.942} & \multicolumn{1}{c|}{0.05} & \multicolumn{1}{c|}{0m 37s} & \multicolumn{1}{c|}{1.163} & \multicolumn{1}{c|}{0.929} & \multicolumn{1}{c|}{0.053} & \multicolumn{1}{c|}{1m 56s} & \multicolumn{1}{c|}{1.215} & \multicolumn{1}{c|}{\textbf{0.961}} & \multicolumn{1}{c|}{0.044} \\ \hline
\rowcolor{Gray}
\multicolumn{1}{|l}{\textit{\textbf{Synthetic}}} & \multicolumn{1}{c}{} & \multicolumn{1}{c}{} & \multicolumn{1}{c}{} & \multicolumn{1}{c}{} & \multicolumn{1}{c}{} & \multicolumn{1}{c}{} & \multicolumn{1}{c}{} & \multicolumn{1}{c}{} & \multicolumn{1}{c}{} & \multicolumn{1}{c}{} & \multicolumn{1}{c}{} & \multicolumn{1}{c|}{} \\ \hline
\multicolumn{1}{|l|}{BlueTiles} & \multicolumn{1}{c|}{0m 28s} & \multicolumn{1}{c|}{1.141} & \multicolumn{1}{c|}{0.792} & \multicolumn{1}{c|}{0.256} & \multicolumn{1}{c|}{0m 23s} & \multicolumn{1}{c|}{1.141} & \multicolumn{1}{c|}{0.816} & \multicolumn{1}{c|}{0.204} & \multicolumn{1}{c|}{0m 52s} & \multicolumn{1}{c|}{1.161} & \multicolumn{1}{c|}{\textbf{0.871}} & \multicolumn{1}{c|}{0.182} \\ \hline
\multicolumn{1}{|l|}{BrickWall} & \multicolumn{1}{c|}{0m 23s} & \multicolumn{1}{c|}{1.094} & \multicolumn{1}{c|}{0.667} & \multicolumn{1}{c|}{0.144} & \multicolumn{1}{c|}{0m 24s} & \multicolumn{1}{c|}{1.098} & \multicolumn{1}{c|}{0.69} & \multicolumn{1}{c|}{0.142} & \multicolumn{1}{c|}{0m 47s} & \multicolumn{1}{c|}{1.1} & \multicolumn{1}{c|}{\textbf{0.703}} & \multicolumn{1}{c|}{0.141} \\ \hline
\multicolumn{1}{|l|}{Vision} & \multicolumn{1}{c|}{29s} & \multicolumn{1}{c|}{1.181} & \multicolumn{1}{c|}{0.938} & \multicolumn{1}{c|}{0.09} & \multicolumn{1}{c|}{23s} & \multicolumn{1}{c|}{1.162} & \multicolumn{1}{c|}{0.916} & \multicolumn{1}{c|}{0.113} & \multicolumn{1}{c|}{52s} & \multicolumn{1}{c|}{1.211} & \multicolumn{1}{c|}{\textbf{0.972}} & \multicolumn{1}{c|}{0.066} \\ \hline
\multicolumn{1}{|l|}{HandWritten} & \multicolumn{1}{c|}{0m 37s} & \multicolumn{1}{c|}{1.123} & \multicolumn{1}{c|}{0.878} & \multicolumn{1}{c|}{0.081} & \multicolumn{1}{c|}{0m 23s} & \multicolumn{1}{c|}{1.131} & \multicolumn{1}{c|}{0.907} & \multicolumn{1}{c|}{0.077} & \multicolumn{1}{c|}{1m 0s} & \multicolumn{1}{c|}{1.156} & \multicolumn{1}{c|}{\textbf{0.938}} & \multicolumn{1}{c|}{0.075} \\ \hline
\rowcolor{Gray}
\multicolumn{1}{|l}{\textit{\textbf{Real2}}} & \multicolumn{1}{c}{} & \multicolumn{1}{c}{} & \multicolumn{1}{c}{} & \multicolumn{1}{c}{} & \multicolumn{1}{c}{} & \multicolumn{1}{c}{} & \multicolumn{1}{c}{} & \multicolumn{1}{c}{} & \multicolumn{1}{c}{} & \multicolumn{1}{c}{} & \multicolumn{1}{c}{} & \multicolumn{1}{c|}{} \\ \hline
\multicolumn{1}{|l|}{Middle} & \multicolumn{1}{c|}{0m 13s} & \multicolumn{1}{c|}{1.192} & \multicolumn{1}{c|}{0.838} & \multicolumn{1}{c|}{0.139} & \multicolumn{1}{c|}{0m 7s} & \multicolumn{1}{c|}{1.211} & \multicolumn{1}{c|}{0.85} & \multicolumn{1}{c|}{0.165} & \multicolumn{1}{c|}{0m 20s} & \multicolumn{1}{c|}{1.23} & \multicolumn{1}{c|}{\textbf{0.914}} & \multicolumn{1}{c|}{0.101} \\ \hline
\multicolumn{1}{|l|}{Small} & \multicolumn{1}{c|}{0m 9s} & \multicolumn{1}{c|}{1.169} & \multicolumn{1}{c|}{0.763} & \multicolumn{1}{c|}{0.164} & \multicolumn{1}{c|}{0m 6s} & \multicolumn{1}{c|}{1.182} & \multicolumn{1}{c|}{0.772} & \multicolumn{1}{c|}{0.206} & \multicolumn{1}{c|}{0m 16s} & \multicolumn{1}{c|}{1.195} & \multicolumn{1}{c|}{\textbf{0.849}} & \multicolumn{1}{c|}{0.133} \\ \hline
\multicolumn{1}{|l|}{Tiny} & \multicolumn{1}{c|}{0m 11s} & \multicolumn{1}{c|}{1.166} & \multicolumn{1}{c|}{0.661} & \multicolumn{1}{c|}{0.201} & \multicolumn{1}{c|}{0m 7s} & \multicolumn{1}{c|}{1.176} & \multicolumn{1}{c|}{0.698} & \multicolumn{1}{c|}{0.263} & \multicolumn{1}{c|}{0m 19s} & \multicolumn{1}{c|}{1.186} & \multicolumn{1}{c|}{\textbf{0.745}} & \multicolumn{1}{c|}{0.19} \\ \hline
 &  &  &  &  &  &  &  &  &  &  &  &  \\ \hline
\multicolumn{1}{|l|}{\textbf{}} & \multicolumn{4}{c|}{\textbf{SBR-RPCA}\cite{Oreifej2011}} & \multicolumn{4}{c|}{\textbf{LWB}\cite{Tian2009}} & \multicolumn{4}{c|}{\textbf{DL}\cite{Li2018}} \\ \hline
\multicolumn{1}{|l|}{\textbf{Dataset}} & \multicolumn{1}{c|}{\textbf{Time}} & \multicolumn{1}{c|}{\textbf{NMI}} & \multicolumn{1}{c|}{\textbf{SSIM}} & \multicolumn{1}{c|}{\textbf{RMSE}} & \multicolumn{1}{c|}{\textbf{Time}} & \multicolumn{1}{c|}{\textbf{NMI}} & \multicolumn{1}{c|}{\textbf{SSIM}} & \multicolumn{1}{c|}{\textbf{RMSE}} & \multicolumn{1}{c|}{\textbf{Time}} & \multicolumn{1}{c|}{\textbf{NMI}} & \multicolumn{1}{c|}{\textbf{SSIM}} & \multicolumn{1}{c|}{\textbf{RMSE}} \\ \hline
\rowcolor{Gray}
\multicolumn{1}{|l}{\textit{\textbf{Real1}}} & \multicolumn{1}{c}{} & \multicolumn{1}{c}{} & \multicolumn{1}{c}{} & \multicolumn{1}{c}{} & \multicolumn{1}{c}{} & \multicolumn{1}{c}{} & \multicolumn{1}{c}{} & \multicolumn{1}{c}{} & \multicolumn{1}{c}{} & \multicolumn{1}{c}{} & \multicolumn{1}{c}{} & \multicolumn{1}{c|}{}
\\ \hline
\multicolumn{1}{|l|}{Cartoon} & \multicolumn{1}{c|}{3h 2m} & \multicolumn{1}{c|}{1.173} & \multicolumn{1}{c|}{0.843} & \multicolumn{1}{c|}{0.111} & \multicolumn{1}{c|}{0h 54m} & \multicolumn{1}{c|}{1.152} & \multicolumn{1}{c|}{0.836} & \multicolumn{1}{c|}{0.095} & \multicolumn{1}{c|}{3s} & \multicolumn{1}{c|}{1.203} & \multicolumn{1}{c|}{0.803} & \multicolumn{1}{c|}{0.162} \\ \hline
\multicolumn{1}{|l|}{Checker} & \multicolumn{1}{c|}{4h 9m} & \multicolumn{1}{c|}{1.158} & \multicolumn{1}{c|}{0.791} & \multicolumn{1}{c|}{0.239} & \multicolumn{1}{c|}{1h 37m} & \multicolumn{1}{c|}{1.105} & \multicolumn{1}{c|}{0.66} & \multicolumn{1}{c|}{0.322} & \multicolumn{1}{c|}{3s} & \multicolumn{1}{c|}{1.129} & \multicolumn{1}{c|}{0.544} & \multicolumn{1}{c|}{0.384} \\ \hline
\multicolumn{1}{|l|}{Dices} & \multicolumn{1}{c|}{3h 58m} & \multicolumn{1}{c|}{1.1} & \multicolumn{1}{c|}{0.758} & \multicolumn{1}{c|}{0.17} & \multicolumn{1}{c|}{1h 26m} & \multicolumn{1}{c|}{1.086} & \multicolumn{1}{c|}{0.783} & \multicolumn{1}{c|}{0.126} & \multicolumn{1}{c|}{3s} & \multicolumn{1}{c|}{1.085} & \multicolumn{1}{c|}{0.637} & \multicolumn{1}{c|}{0.242} \\ \hline
\multicolumn{1}{|l|}{Bricks} & \multicolumn{1}{c|}{3h 43m} & \multicolumn{1}{c|}{1.128} & \multicolumn{1}{c|}{0.686} & \multicolumn{1}{c|}{0.192} & \multicolumn{1}{c|}{1h 24m} & \multicolumn{1}{c|}{1.118} & \multicolumn{1}{c|}{0.673} & \multicolumn{1}{c|}{0.225} & \multicolumn{1}{c|}{3s} & \multicolumn{1}{c|}{1.058} & \multicolumn{1}{c|}{0.49} & \multicolumn{1}{c|}{0.422} \\ \hline
\multicolumn{1}{|l|}{Elephant} & \multicolumn{1}{c|}{3h 7m} & \multicolumn{1}{c|}{1.075} & \multicolumn{1}{c|}{0.516} & \multicolumn{1}{c|}{0.257} & \multicolumn{1}{c|}{0h 59m} & \multicolumn{1}{c|}{1.068} & \multicolumn{1}{c|}{0.584} & \multicolumn{1}{c|}{0.204} & \multicolumn{1}{c|}{3s} & \multicolumn{1}{c|}{1.075} & \multicolumn{1}{c|}{0.378} & \multicolumn{1}{c|}{0.347} \\ \hline
\multicolumn{1}{|l|}{Eye} & \multicolumn{1}{c|}{4h 4m} & \multicolumn{1}{c|}{1.179} & \multicolumn{1}{c|}{0.913} & \multicolumn{1}{c|}{0.104} & \multicolumn{1}{c|}{1h 22m} & \multicolumn{1}{c|}{1.155} & \multicolumn{1}{c|}{0.903} & \multicolumn{1}{c|}{0.089} & \multicolumn{1}{c|}{3s} & \multicolumn{1}{c|}{1.141} & \multicolumn{1}{c|}{0.804} & \multicolumn{1}{c|}{0.191} \\ \hline
\multicolumn{1}{|l|}{Math} & \multicolumn{1}{c|}{4h 34m} & \multicolumn{1}{c|}{1.1} & \multicolumn{1}{c|}{0.841} & \multicolumn{1}{c|}{0.102} & \multicolumn{1}{c|}{3h 0m} & \multicolumn{1}{c|}{1.067} & \multicolumn{1}{c|}{0.766} & \multicolumn{1}{c|}{0.1} & \multicolumn{1}{c|}{3s} & \multicolumn{1}{c|}{1.073} & \multicolumn{1}{c|}{0.678} & \multicolumn{1}{c|}{0.139} \\ \hline
\rowcolor{Gray}
\multicolumn{1}{|l}{\textit{\textbf{Synthetic}}} & \multicolumn{1}{c}{} & \multicolumn{1}{c}{} & \multicolumn{1}{c}{} & \multicolumn{1}{c}{} & \multicolumn{1}{c}{} & \multicolumn{1}{c}{} & \multicolumn{1}{c}{} & \multicolumn{1}{c}{} & \multicolumn{1}{c}{} & \multicolumn{1}{c}{} & \multicolumn{1}{c}{} & \multicolumn{1}{c|}{} \\ \hline
\multicolumn{1}{|l|}{BlueTiles} & \multicolumn{1}{c|}{2h 5m} & \multicolumn{1}{c|}{1.142} & \multicolumn{1}{c|}{0.763} & \multicolumn{1}{c|}{0.372} & \multicolumn{1}{c|}{0h 55m} & \multicolumn{1}{c|}{1.104} & \multicolumn{1}{c|}{0.72} & \multicolumn{1}{c|}{0.204} & \multicolumn{1}{c|}{3s} & \multicolumn{1}{c|}{1.091} & \multicolumn{1}{c|}{0.365} & \multicolumn{1}{c|}{1.067} \\ \hline
\multicolumn{1}{|l|}{BrickWall} & \multicolumn{1}{c|}{2h 30m} & \multicolumn{1}{c|}{1.093} & \multicolumn{1}{c|}{0.666} & \multicolumn{1}{c|}{0.158} & \multicolumn{1}{c|}{1h 0m} & \multicolumn{1}{c|}{1.066} & \multicolumn{1}{c|}{0.481} & \multicolumn{1}{c|}{0.19} & \multicolumn{1}{c|}{3s} & \multicolumn{1}{c|}{1.079} & \multicolumn{1}{c|}{0.479} & \multicolumn{1}{c|}{0.218} \\ \hline
\multicolumn{1}{|l|}{Vision} & \multicolumn{1}{c|}{3h 4} & \multicolumn{1}{c|}{1.115} & \multicolumn{1}{c|}{0.739} & \multicolumn{1}{c|}{0.216} & \multicolumn{1}{c|}{36m} & \multicolumn{1}{c|}{1.021} & \multicolumn{1}{c|}{0.446} & \multicolumn{1}{c|}{0.266} & \multicolumn{1}{c|}{3s} & \multicolumn{1}{c|}{1.095} & \multicolumn{1}{c|}{0.599} & \multicolumn{1}{c|}{0.215} \\ \hline
\multicolumn{1}{|l|}{HandWritten} & \multicolumn{1}{c|}{0h 0m} & \multicolumn{1}{c|}{1.112} & \multicolumn{1}{c|}{0.851} & \multicolumn{1}{c|}{0.12} & \multicolumn{1}{c|}{0h 52m} & \multicolumn{1}{c|}{1.073} & \multicolumn{1}{c|}{0.678} & \multicolumn{1}{c|}{0.147} & \multicolumn{1}{c|}{3s} & \multicolumn{1}{c|}{1.074} & \multicolumn{1}{c|}{0.546} & \multicolumn{1}{c|}{0.177} \\ \hline
\rowcolor{Gray}
\multicolumn{1}{|l}{\textit{\textbf{Real2}}} & \multicolumn{1}{c}{} & \multicolumn{1}{c}{} & \multicolumn{1}{c}{} & \multicolumn{1}{c}{} & \multicolumn{1}{c}{} & \multicolumn{1}{c}{} & \multicolumn{1}{c}{} & \multicolumn{1}{c}{} & \multicolumn{1}{c}{} & \multicolumn{1}{c}{} & \multicolumn{1}{c}{} & \multicolumn{1}{c|}{}
\\ \hline
\multicolumn{1}{|l|}{Middle} & \multicolumn{1}{c|}{1h 28m} & \multicolumn{1}{c|}{1.189} & \multicolumn{1}{c|}{0.782} & \multicolumn{1}{c|}{0.204} & \multicolumn{1}{c|}{0h 54m} & \multicolumn{1}{c|}{1.163} & \multicolumn{1}{c|}{0.761} & \multicolumn{1}{c|}{0.194} & \multicolumn{1}{c|}{3s} & \multicolumn{1}{c|}{1.122} & \multicolumn{1}{c|}{0.512} & \multicolumn{1}{c|}{0.307} \\ \hline
\multicolumn{1}{|l|}{Small} & \multicolumn{1}{c|}{1h 21m} & \multicolumn{1}{c|}{1.153} & \multicolumn{1}{c|}{0.741} & \multicolumn{1}{c|}{0.181} & \multicolumn{1}{c|}{0h 33m} & \multicolumn{1}{c|}{1.151} & \multicolumn{1}{c|}{0.688} & \multicolumn{1}{c|}{0.198} & \multicolumn{1}{c|}{3s} & \multicolumn{1}{c|}{1.114} & \multicolumn{1}{c|}{0.418} & \multicolumn{1}{c|}{0.323} \\ \hline
\multicolumn{1}{|l|}{Tiny} & \multicolumn{1}{c|}{1h 6m} & \multicolumn{1}{c|}{1.161} & \multicolumn{1}{c|}{0.657} & \multicolumn{1}{c|}{0.395} & \multicolumn{1}{c|}{0h 34m} & \multicolumn{1}{c|}{1.167} & \multicolumn{1}{c|}{0.654} & \multicolumn{1}{c|}{0.238} & \multicolumn{1}{c|}{3s} & \multicolumn{1}{c|}{1.144} & \multicolumn{1}{c|}{0.492} & \multicolumn{1}{c|}{0.306} \\ \hline
 & \multicolumn{1}{l}{} & \multicolumn{1}{l}{} & \multicolumn{1}{l}{} & \multicolumn{1}{l}{} & \multicolumn{1}{l}{} & \multicolumn{1}{l}{} & \multicolumn{1}{l}{} & \multicolumn{1}{l}{} & \multicolumn{1}{l}{} & \multicolumn{1}{l}{} & \multicolumn{1}{l}{} & \multicolumn{1}{l}{}
\end{tabular}
\label{tab:results}
\end{table*}
Additionally, we observed that SBR-RPCA, DL and LWB \emph{did not preserve the image structure} as well as our method (see grid-lines in `Elephant', the words `Fluctuation' or `Distortion' in `Middle' and alphabets E, T, Z, large D in `Eye' in Fig. \ref{fig:comparison_results}. We do believe the DL method \cite{Li2018} may yield improved results if their network were trained to restore multiple frames together, as opposed to single frames individually (as done by their current algorithm) - which ignores the temporal aspect leading to loss in performance. All in all, our results show that exploiting spatio-temporal properties of water waves for this task is indeed useful. 
\\
\textbf{Computational Time:} The compute time for all methods (measured on a 2.6GHz Intel Xeon machine with 32GB RAM) are presented in Table \ref{tab:results}. The DL method is the fastest, whereas our methods are much faster than SBR-RPCA and LWB. However for CS, the YALL1 solver uses GPU support, which is unavailable in the authors' code for SBR-PCA and LWB. We note that although cross-validation is an excellent way to pick the $\hat{\lambda}$ parameter in Eqn. \ref{eq:lasso}, we found that the optimal choice of this parameter did not change much across datasets. Also small changes in $\hat{\lambda}$ did not affect the performance much. Hence the time for cross-validation is not included in Table \ref{tab:results}.
\\\textbf{Verification of Fourier Sparsity:} Here, we demonstrate the sparsity of the MVFs from real underwater sequences. This is shown in Fig. \ref{fig:DT_fourier_sparsity} for the `Elephant' sequence (similar plots can be generated for all other sequences). We note that the actual MVF can only be \emph{estimated}. However, we contend that the MVF estimated by our CS method is a good approximation to the actual one. This is evident when comparing the quality of the estimated mean image with the original undistorted image (gathered in still water). For further quantification, we also tracked the same $N$ points (from the distorted video $I_d$) that were used for the CS algorithm in Sec. \ref{subsubsec:mvf_cs}, in the restored video ($I_r$) produced by the CS step. This gave us new DTs $\{\boldsymbol{\hat{d}_i}\}_{i=1}^N$. We computed a measure of the motion reduction given as $MR \triangleq \textrm{median}_{i \in \{1,...,N\}} \frac{\|\boldsymbol{\hat{d_i}}-\boldsymbol{d_i}\|_2}{\|\boldsymbol{d_i}\|_2}$. We note that in most cases, we achieve more than $90\%$ motion reduction by the CS step - see Table \ref{tab:motion_reduction}. We have included a few videos in the \texttt{supplemental material} for the visual comparison of the estimated MVF w.r.t. the ground truth MVF.

\begin{table}[]
    \centering
    \begin{tabular}{|l|l|l|l|}
\hline   
\rowcolor{Gray}
Dataset &	$N$ & $MR$ & $\sigma_{motion}$\\
\hline
Cartoon (Real1)	& 1029	& 94.11\% & 7.42 \\
Checker (Real1)	& 3149	& 85.25\% & 8.5 \\
Dices (Real1)	& 2230	& 91.9\% & 7.75 \\
Bricks (Real1)	& 1300	& 87.38\% & 7.42 \\
Elephant (Real1) & 3670	& 97.7\% & 7.34 \\
Eye (Real1)	& 1647	& 81.66\% & 7.84 \\
Math (Real1)	& 2309	& 96.12\% & 5.64 \\ 
\hline
BlueTiles (Synth.)	& 2192	& 94.77\% & 5.71 \\
BrickWall (Synth.)	& 3134	& 94.57\% & 8.68\\
Vision (Synth.) & 5266	& 93.49\% & 6.77  \\
HandWritten	(Synth.) & 3789	& 95.82\% & 4.33 \\ \hline
Middle (Real2)	& 785	& 96.34\% & 5.65 \\ Small (Real2)	& 993	& 97.26\% & 4.22 \\ 
Tiny (Real2)	& 155	& 87.84\% & 5.03\\
\hline
    \end{tabular}
    \caption{$\#$salient points $N$,  motion reduction $MR$, and $\sigma_{motion}$ for different videos}
    \label{tab:motion_reduction}
\end{table}

\textbf{Effect of number of frames $T$:}
In the absence of attenuation, a large $T$ helps improve the performance of our algorithm, due to better Fourier sparsity and better approximation of the COT. In practice, we observed on real datasets that just 100 frames were sufficient to yield good reconstruction results. Further increase in $T$ had an insignificant impact on the result quality. A graph showing the effect of $T$ on reconstruction from real sequences is shown in Fig. \ref{fig:NT}. 
\\\textbf{Effect of number of tracked points $N$:} The number and accuracy of DTs affects the performance of the CS algorithm. The number of DTs varied across datasets, depending on the number of available salient feature points, but was always less than $0.03N_xN_y$. The slightly lower performance of CS on the `Tiny' sequence for example (see Table \ref{tab:results}) is due to the small number of available salient points, less than $0.002N_xN_y$ - see Table \ref{tab:motion_reduction}. A graph showing the positive impact of increase in the number of good quality tracks (upto a point, beyond which the performance saturates) is shown in Fig. \ref{fig:NT}. We note, that we have ensured good quality of the trajectories for further stages of our algorithms, as mentioned in Sec. \ref{subsubsec:tracking}. We considered \emph{global sparsity in this work, as opposed to sparsity of small spatial or spatio-temporal patches}. Since, many patches without any salient points could exist.

\section{Conclusion}
\label{sec:disc}
We have presented two methods for correction of refractive deformations due to a wavy water surface, one based on a novel application of CS for interpolating MVFs starting from a small set of salient PTs (and their DTs), and the other based on polynomial image expansions. In both cases, we obtain results superior to the state of the art at low computational cost. Avenues for future work include (1) extending the CS algorithm to handle moving objects; (2) studying the effect of depth variation, perspective projection or wave attenuation on the results of our algorithms; and (3) exploring MVF sparsity in other bases instead of the DFT.
\\\textbf{Acknowledgements}:  The authors wish to thank the Qualcomm Innovation Fellowship Program (India) for supporting this work, NVIDIA Corp. for donation of a Titan Xp GPU, and Prof. Manasa Ranjan Behera of the Civil Engineering Department at IITB, for the wave-flume facility to acquire real data.\\\textbf{The source code, datasets and supplemental material}  can be accessed at \cite{GitRepo}, \cite{ProjectPage}.

\clearpage
{\small
\bibliographystyle{ieee}
\bibliography{ms}
}
\end{document}